\pdfoutput=1
\documentclass[11pt]{article}

\usepackage{acl}

\usepackage{times}
\usepackage{latexsym}                                                         
\usepackage[T1]{fontenc}
\usepackage[utf8]{inputenc}
\usepackage{microtype}
\usepackage{inconsolata}
\usepackage{graphicx}
\usepackage{amsmath}
\usepackage{makecell}

\usepackage{multirow}
\usepackage{booktabs}

\usepackage{times}
\usepackage{latexsym}

\usepackage{multicol}
\usepackage{xcolor}
\usepackage{mathtools}
\usepackage{enumitem}
\usepackage[noend]{algpseudocode}
\usepackage{subcaption}
\usepackage{color}
\usepackage{colortbl}
\usepackage{amssymb}
\usepackage{pifont}
\usepackage{bbm}
\usepackage{svg}

\title{Mentor-KD: Making Small Language Models Better Multi-step Reasoners}

\author{
    \textbf{Hojae Lee}\textsuperscript{1}\thanks{\enspace These authors contributed equally to this work.}, \
    \textbf{Junho Kim}\textsuperscript{2}\footnotemark[1], \
    \textbf{SangKeun Lee}\textsuperscript{1,2} \\
  \textsuperscript{1}Department of Computer Science and Engineering \
  \textsuperscript{2}Department of Artificial Intelligence \\  
  Korea University, Seoul, Republic of Korea\\
  \texttt{\{22leehojae, monocrat, yalphy\}@korea.ac.kr} \\
}

\begin{document}
\maketitle

\begin{abstract}
Large Language Models (LLMs) have displayed remarkable performances across various complex tasks by leveraging Chain-of-Thought (CoT) prompting. Recently, studies have proposed a Knowledge Distillation (KD) approach, \textit{reasoning distillation}, which transfers such reasoning ability of LLMs through fine-tuning language models of multi-step rationales generated by LLM teachers.
However, they have inadequately considered two challenges regarding insufficient distillation sets from the LLM teacher model, in terms of 1) data quality and 2) soft label provision. 
In this paper, we propose Mentor-KD, which effectively distills the multi-step reasoning capability of LLMs to smaller LMs while addressing the aforementioned challenges. 
Specifically, we exploit a mentor, intermediate-sized task-specific fine-tuned model, to augment additional CoT annotations and provide soft labels for the student model during reasoning distillation. We conduct extensive experiments and confirm Mentor-KD's effectiveness across various models and complex reasoning tasks\footnote{Our code and data are available at \url{https://github.com/2hojae/mentor-kd}}.

%developed methods to transfer such multi-step reasoning abilities from black-box LLMs to open-sourced smaller language models (LMs) through Knowledge Distillation (KD). 

\end{abstract}

\section{Introduction} \label{sec:introduction}

% LMs have shown remarkable in-context learning performance in various downstream tasks, as model size is scaled[1]. However, LLMs struggle from complex tasks that require multistep reasoning, such as commonsense and arithmetic tasks[2].

% limited data scale from LLM models may lead student be over-fitted on reasonings. -> 다른 과적합 문제도 적용해야 할수도?

%Large Language Models (LLMs) have shown impressive emergent capabilities, showcasing their competence in various natural language processing (NLP) tasks. 

Large Language Models (LLMs) have shown impressive emergent capabilities, showing their competence on a variety of reasoning tasks in the Natural Language Processing (NLP) landscape \cite{brown-etal-2020-gpt3, rae-etal-2021-gopher, hoffmann-etal-2022-chinchilla, palm}. One particularly interesting strategy for this approach is Chain-of-Thought (CoT) prompting, which elicits multi-step reasoning abilities of LLMs by explicitly generating intermediate reasoning steps for complex tasks \cite{wei-etal-2022-cot}. However, such reasoning abilities have been shown to only manifest in language models (LMs) with over hundreds of billion parameters \cite{chung-etal-2022-flant5, wei-etal-2022-emergent}, which require significant computational resources or expensive API calls, restricting their deployment on resource-limited scenarios.

To circumvent these deployment challenges, previous works \cite{ho-etal-2023-ftcot, li-etal-2023-sctod, magister-etal-2023-teaching} have followed a knowledge distillation (KD) approach, \textit{reasoning distillation}, which transfers the multi-step reasoning ability of LLMs to small LMs.
%Knowledge Distillation (KD) has been proposed to transfer the reasoning capability of LLMs to small LMs \cite{ho-etal-2023-ftcot, li-etal-2023-sctod, magister-etal-2023-teaching}. 
%by teaching them how to solve complex tasks 
The KD pipeline generally applies In-Context Learning (ICL) on the LLM teacher model to generate outputs (e.g., multi-step rationales) as distillation sets, and then utilizes them to fine-tune the student model. Previous studies have shown that reasoning distillation can significantly improve student performances and may even outperform their LLM teachers on specific tasks \cite{ho-etal-2023-ftcot, chen-etal-2023-mcckd}.

%To circumvent these deployment challenges, recent studies have proposed reasoning distillation, which transfers the reasoning capability of LLMs to small language models by teaching how to solve complex tasks in a step-by-step manner \cite{ho-etal-2023-ftcot, li-etal-2023-sctod, magister-etal-2023-teaching}. Specifically, the student model acquires the reasoning ability of LLMs by learning multi-step rationales generated by the LLM teacher. Previous studies have shown that the performances of the student can be significantly improved, and may even outperform their teachers \cite{ho-etal-2023-ftcot} by leveraging reasoning distillation.

\begin{figure*}[t]
\centering
    \includegraphics[width=\textwidth]{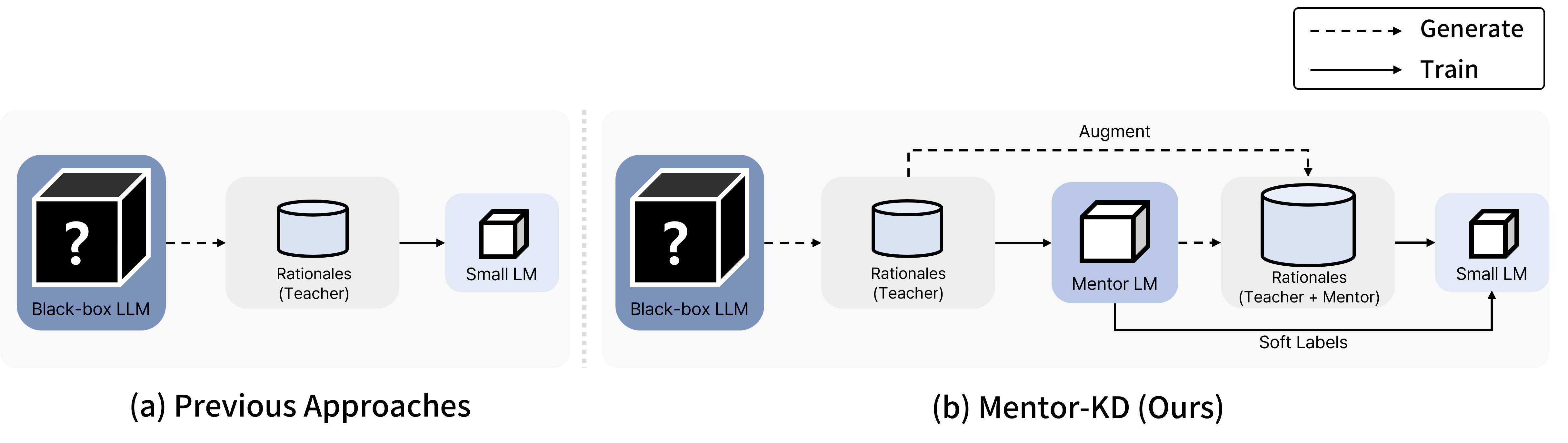}
\caption{Comparison between (a) previous approaches of reasoning distillation and (b) Mentor-KD (ours). Our framework utilizes an intermediate-sized task-specific mentor model to complement the distillation sets of teachers.}
\label{Figure1}
\end{figure*}

However, previous approaches to reasoning distillation have two challenges arising from insufficient distillation sets generated by LLM teachers. 
First, as LLMs may not have access to task-specific data, the quality of the rationales for distillation can be low (e.g., only 58\% accuracy on GPT-3.5 for StrategyQA). 
The low quality of LLM teacher rationales limits the number of reasoning rationales to only a small set of correct ones due to the exclusion of incorrect rationales that negatively affect student performances \cite{ho-etal-2023-ftcot}.
%These incorrect rationales from teachers negatively affect student performance \cite{ho-etal-2023-ftcot} by limiting the number of reasoning rationales to only a small set of correct ones. 
% ------ 이전 라이팅 ------%
% Regarding excluding incorrect teacher rationales, which negatively affects student performance \cite{ho-etal-2023-ftcot}, the low quality of LLM teacher rationales limits the number of reasoning rationales to only a small set of correct ones.
% ----------------------- %
%Given that including incorrect teacher rationales would negatively affect student performance \cite{ho-etal-2023-ftcot}, the low-quality LLM teacher rationales limit the number of reasoning rationales to only a small set of correct ones.
% Second, due to the restricted accessibility of black-box LLM teachers, 
Second, because accessibility of black-box LLM teachers is generally restricted, the student model cannot mimic the predictive behavior and knowledge from the soft labels \cite{hinton-etal-2015-kd}. 
Such oversights may lead to the student model being over-fitted on limited distillation sets from teacher models and undermine its generalization capabilities.
To address these challenges, we propose Mentor-KD, a novel reasoning distillation framework that effectively distills the multi-step reasoning capability of LLMs. 
%Motivated by the task-specific small LM achieves comparable performance with LLM teachers \cite{}, we introduce a task-specific mentor model, which complements the knowledge of the LLM teacher during reasoning distillation.
Our core idea is to introduce a mentor, an intermediate-sized task-specific model, that complements the LLM teacher's knowledge during reasoning distillation.
To this end, we first fine-tune the mentor models on specific tasks and generate both CoT rationales and soft labels to augment distillation sets. %for training the student. 
By leveraging task-specific mentors whose power is concentrated toward a specific target ability, Mentor-KD effectively addresses two issues through training on more diverse rationales and intrinsic knowledge from soft labels.
%With the help of task-specific mentors, 
%Specifically, the mentor model addresses the aforementioned two issues by augmenting distillation sets of both CoT rationales and soft labels for training the student.
%By leveraging task-specific mentor models whose powers are concentrated towards a specific target ability, Mentor-KD enables student models to be trained on more diverse rationales and on intrinsic knowledge. 
%that such task-specific models possess.
% powers are concentrated 워딩은 (Fu et al., 2023)에서 가져옴.

% By leveraging the task-specific models as mentors, Mentor-KD enables student models to be trained on more diverse CoT rationales and intrinsic knowledge in soft labels from mentors.
%that the mentor model internally possesses, ultimately boosting the student model's reasoning ability.
%By leveraging the mentor model which has comparable performance with LLMs on specific tasks, Mentor-KD enables student models to be trained on more diverse CoT rationales, as well as on the intrinsic knowledge in soft labels that the mentor model internally possesses, ultimately boosting the student model's reasoning ability.
%task specific 한 학생 모델이 교사와 견줄만한 성능을 가질 수 있다는 사실에서 동기를 받아, 우리는 교사-학생 중간 크기의 task specific fine-tuned 된 mentor 모델을 소개하여 교사의 지식을 보충한다. 구체적으로, 

We conduct extensive experiments on various types of complex reasoning tasks, including commonsense, arithmetic, logical, and symbolic reasoning tasks. The experimental results clearly demonstrate the superiority of our method over baselines leveraging knowledge only from LLMs. 
In addition, we verify that the mentor model can generate a substantial number of correct reasoning samples compared to other LLM baselines, highlighting the effectiveness of our method as means of data augmentation.
%our method's effectiveness as a means of data augmentation. 
%Lastly, we demonstrate that our Mentor-KD successfully generates more diverse reasoning datasets for training the student models. 
Lastly, we demonstrate that our Mentor-KD significantly improves student performances in low-resource scenarios, indicating its cost-efficiency.
In summary, the contributions of this paper include the following:

\begin{itemize}
    \item {We propose Mentor-KD, a novel reasoning distillation framework, which improves the reasoning ability of small LMs considering the limitations of insufficient distillation sets from LLM teachers.}
    %\item {We present a mentor network to complement the limited LLM teachers training datasets by generating both additional reasoning samples and soft labels}
    \item {We introduce a mentor model to additionally generate both rationale samples and soft labels to complement the limited training datasets from the LLM teachers.}    
    \item {We demonstrate that Mentor-KD improves the effectiveness of reasoning distillation on students with various types of reasoning and models through extensive experiments.}
\end{itemize}

\begin{figure*}[t]
\centering
    \includegraphics[width=\textwidth]{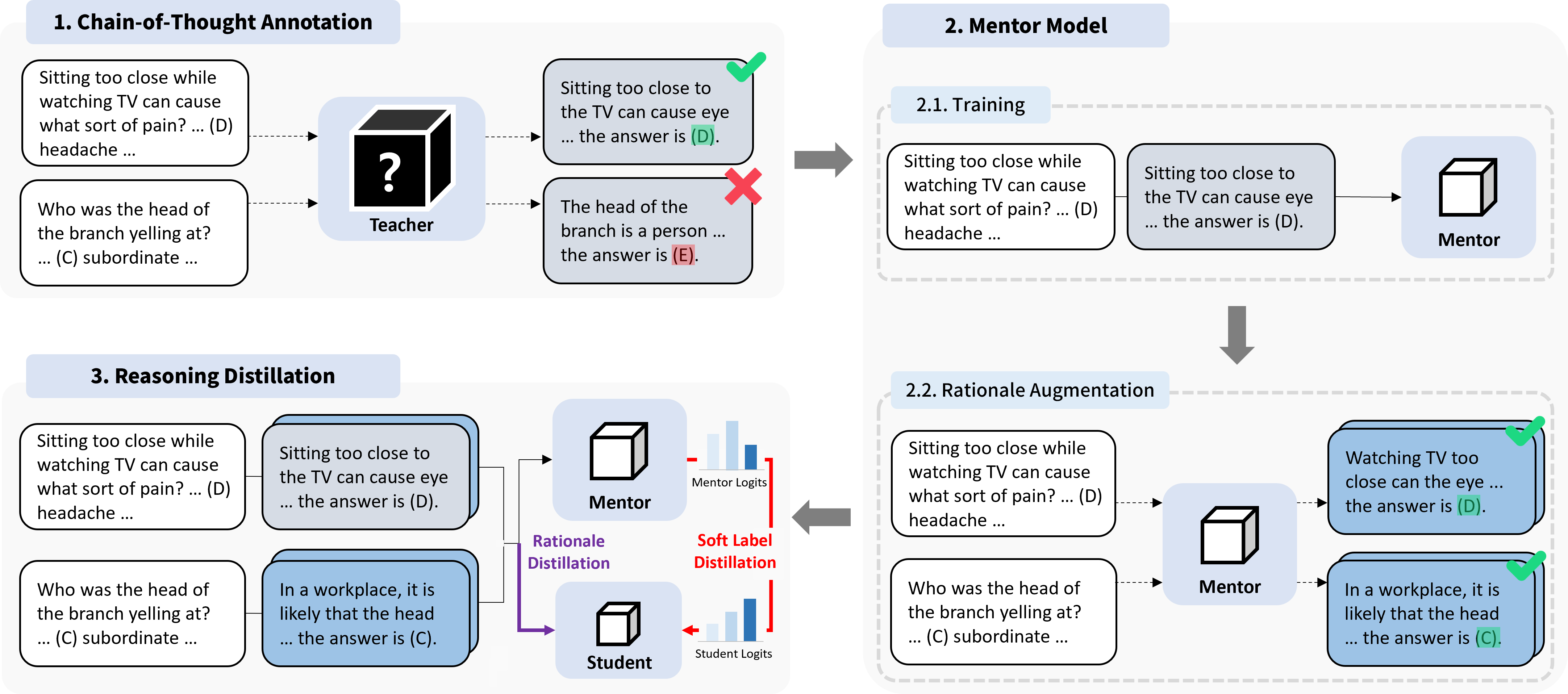}
\caption{A general overview of our proposed framework, Mentor-KD. Mentor-KD is composed of three steps. First, CoT annotations are initially collected from the teacher LLM and filtered. Second, the preserved annotations are used to train the mentor model, and the trained mentor model augments multi-step rationales. Lastly, the student model is trained on annotations from the teacher and the student, as well as soft labels from the mentor model.}
\label{fig:main_figure}
\end{figure*}

% A general overview of our proposed framework, Mentor-KD. Mentor-KD is composed of three steps. First, CoT annotations are initially collected from the teacher LLM and filtered. Second, the preserved annotations are used to train the mentor model, and the trained mentor model augments the multi-step rationales, including those where the teacher was incorrect. Lastly, the annotations from both the teacher and the mentor are utilized for training the student model, where soft labels from the mentor model are also incorporated during training.

\section{Related Works}

\subsection{Chain-of-Thought Prompting}

% 수정완료료
CoT prompting is a method that elicits multi-step reasoning abilities of LMs through ICL \cite{wei-etal-2022-cot}. The essence of CoT is that it acts as a guidance of logical progression for LMs to decompose and solve complex reasoning tasks \cite{xia-etal-2024-beyond}. Consequently, it allowed LMs to excel in complex reasoning tasks \cite{kojima-etal-2022-zscot, wang-etal-2023-self_consistency, zhang-etal-2023-autocot} which traditional few-shot learning methods have struggled with \cite{rae-etal-2021-gopher}. Recent works take a step further to improve CoT prompting through enhancing the quality of reasoning steps. \citet{madaan-etal-2023-self_refine} had LMs to iteratively self-refine reasoning through self-feedback, while \citet{gou-etal-2024-critic} leveraged external tools for obtaining feedback. \citet{trivedi-etal-2023-ircot, zhao-etal-2023-vae} incorporated information retrieval systems to enhance the facticity of LMs' reasoning.

Despite the success, previous works \cite{hoffmann-etal-2022-chinchilla, wei-etal-2022-cot, chu-etal-2024-navigate} reported that the merits of reasoning on CoT prompting emerge when LMs are scaled to hundreds of billions of parameters. To address such problems, our work focuses on enabling CoT reasoning to small-scaled LMs through reasoning distillation. 

% Chain-of-Thought prompting is a method to have LMs explicitly generate intermediate solutions prior to arriving at the final prediction. It has successfully enhanced LM's performances across various complex tasks in both few-shot\ cite{wei-etal-2022-cot} and zero=shot \cite{kojima-etal-2022-zscot} scenarios.

% The primary drawback of these methods lies in their failure to fully leverage all the information that can be provided by the teacher. (Towards Cross-Tokenizer Distillation: the Universal Logit Distillation Loss for LLMs, arXiv 2024)

%\subsection{Chain-of-Thought Distillation}
\subsection{Knowledge Distillation for LLMs}
KD \cite{hinton-etal-2015-kd} has been proven to be a promising approach to compress LMs by transferring the predictive behavior (e.g., soft labels) or internal knowledge (e.g., hidden representations) from larger LMs to smaller ones.
%, which allowed students to retain high performances and generalization abilities.
%Hinton, distillBERT(GPT도 포함되므로) representation -> minilm, tutor, gpt 계열 지식증류 하나
However, existing KD methods for pre-trained LMs, which involve distilling the soft labels \cite{distilbert, minillm} or representations \cite{minilm, minilmv2, tutor}, require access to the internal parameters of teachers. 
These requirements pose a significant challenge for leveraging LLMs in KD, regarding their black-box nature and impracticality.
%posing significant challenges for LLMs.
%are usually impractical for LLMs
%posing significant challenges for large language models (LLMs).
%often unfeasible for LLMs.
% internal parameters of the teacher model, which are usually impractical for LLMs.
%However, KD has been difficult to apply to scenarios where the teacher model is a black-box LLM, as access to its soft labels is restricted.

%다시 쓸것. (SLM 은 정의 안되어 있는데 들어가있음)--> 내용 구성다시할 예정
% In turn, recent works have proposed reasoning distillation, which enables smaller LMs to practice multi-step reasoning similar to LLMs by utilizing the rationales generated by LLM teachers, instead of soft labels. The core idea of reasoning distillation follows the paradigm of symbolic knowledge distillation \cite{west-etal-2022-symbolic} which is to fine-tune student models on generations that larger LMs have produced. Motivated by this, \citet{ho-etal-2023-ftcot, magister-etal-2023-teaching, li-etal-2023-sctod, shridhar-etal-2023-socratic_cot} directly fine-tuned student models on CoT rationales that LLMs have generated. \citet{wang-etal-2023-democratizing} had the teacher model provide real-time feedback generations specifically tailored to the student's generations during training. \citet{zhu-etal-2024-distilling} incorporated forms of reasoning chains in natural language and code from the LLM. %However, these works mainly rely on the generations that the teacher model has made and limit the teacher model's distillation signals to such an extent.

In turn, recent works practiced reasoning distillation, which enabled smaller LMs (students) to carry out multi-step reasoning similar to LLMs by utilizing rationales generated by LLM teachers instead of soft labels. For example, \citet{ho-etal-2023-ftcot, magister-etal-2023-teaching, li-etal-2023-sctod} fine-tuned students on multi-step rationales that LLMs generated. Similarly, \citet{shridhar-etal-2023-socratic_cot} had students learn how to decompose a complex question through having LLMs to generate sub-problems to the original question. \citet{wang-etal-2023-democratizing} iteratively employed LLMs to provide real-time feedback specifically tailored to the student's generations. \citet{kang-etal-2023-kard, zhao-etal-2024-prr} leveraged information retrieval systems to enhance the facticity of student's reasoning on knowledge-intensive tasks. Recently, \citet{zhu-etal-2024-pad, zhu-etal-2024-distilling} incorporated multi-step rationales in a code format generated from the LLMs to improve the student's arithmetic reasoning skills. Contemporaneous to our work, \cite{zhou-ai-2024-teaching} also utilized intermediate-sized models for LLM distillation. Our work differs in that we use intermediate-sized models for complementing the teacher model's distillation signals, rather than for filtering the annotations.

% 추가예정
% A concurrent work \cite{zhou-ai-2024-teaching} used intermediate-sized models during LLM distillation. However, our work 

% Similarly, \citet{shridhar-etal-2023-socratic_cot} had students learn how to decompose a complex question through having LLMs to generate sub-problems to the original question

While most previous works have been conducted to improve reasoning distillation by utilizing distillation sets provided by LLMs, we posit that they may be insufficient and may undermine the student's capabilities. In this sense, our work is different in that we complement such insufficiency of LLM teachers.
%by addressing both data quality and soft label provision, ultimately increasing the effectiveness of reasoning distillation.

% Most previous works have been conducted to improve reasoning distillation by utilizing the distillation signals provided by LLMs. However, the abilities of LLMs are spread over multiple dimensions, and small models that are trained on a specific task may work a

% While most previous works have been conducted for improving reasoning distillation by transferring informative rationales generated from LLMs, studies on flawed teacher annotations have not been well explored. Different from the previous works, our work focuses on complementing the insufficient distillation sets from LLM teachers, enabling the student model to be trained in more diverse data samples.
%as well as the internal knowledge that larger models possess.

%we introduce task-specific mentor models to complement the insufficient distillation sets from the teacher, including both CoT rationales and soft labels. 

%In contrast, a key distinction from previous works is that we introduce a task-specific mentor model during reasoning distillation, to complement the insufficient distillation signals from the teacher, ultimately enabling the student model to be trained on more diverse data samples, as well as the internal knowledge that larger models possess.

\section{Methodology}
% on our proposed framework, Mentor-KD, with detailed implementations. 
We elaborate on the detailed implementations of our Mentor-KD.
The core idea is to augment the distillation training set by leveraging a task-specific intermediate-sized mentor model. To this end, we first generate CoT annotations from LLM teacher models (Section~\ref{sec:cot_annotation}). We then fine-tune the mentor model with the distillation set from the LLM teacher, and the trained mentor model generates additional training sets, including both rationales and soft labels (Section~\ref{sec:mentor}). By augmenting both signals from the mentor, we distill the knowledge to student models (Section~\ref{sec:reasoning_distillation}). Figure \ref{fig:main_figure} illustrates an overview of our framework.

%to interleave a mentor model in between the teacher and the student during reasoning distillation.

\subsection{Chain-of-Thought Annotations} \label{sec:cot_annotation}
We use the LLM to obtain CoT annotations composed of a rationale and a final prediction to a question via Zero-shot-CoT \cite{kojima-etal-2022-zscot}. It is a two-staged strategy consisting of reasoning and answer extraction stages, and thus, we induce the LLM to generate a CoT rationale first and subsequently a final prediction afterwards.

Specifically, we first append ``Let's think step by step'' to the question and prompt the LLM to obtain the rationale. In sequence, we prompt the LLM again by incorporating the previously obtained rationale to induce its final prediction. Formally, from a dataset $\mathcal{D} = \{q_i, y_i\}$ where $q_i$ denotes a question and $y_i$ denotes a golden label, our goal is to induce the LLM to generate a step-by-step rationale $r^t_i$ and a final prediction $\hat{y}^t_i$, given $q_i$ as an input. The prompting template takes the form of: "Q: $\{q_i\}$. A: Let's think step by step. $\{r^t_i\}$. Therefore, the answer is $\{\hat{y}^t_i\}$".

%교사가 틀릴 수 있다는 전제를 넣어서 필터링의 필요성을 강조
Afterward, we filter the annotations generated by the LLM. Following previous works \cite{li-etal-2023-sctod, magister-etal-2023-teaching, fu-etal-2023-specializing, lee-etal-2024-lm_guided_cot}, we preserve annotations where the final prediction $\hat{y}^t_i$ matches the golden answer $y_i$ of a sample. Then, the annotations are reformatted into a question-label format for training mentor and student models. More formally, for all annotations $i$ where $\hat{y}^t_i = y_i$, we reformat a data sample $(q_i, r^t_i, \hat{y}^t_i, y_i)$ into $(q_i, l^t_i, y_i)$, where $l^t_i$ takes the form of ``\{$r^t_i$\}. Therefore, the answer is \{$y_i$\}.'' Consequently, we finally construct $\mathcal{D}_{\text{teacher}}=\{(q_i, l^t_i, y_i)\}_{i=1}^N$.

\subsection{Mentor Model} \label{sec:mentor}
%레퍼런스 붙여서 구체적이게 보이기

Here, we describe how our mentor models are trained to concentrate their powers to a specific task, and utilized to complement the insufficient distillation sets of LLM teachers.

% Distillation sets from LLM teachers provide supervision for multi-step reasoning, but they are insufficient due to their low quality on specific tasks as well as their black-box nature. Thus, we introduce an open-sourced intermediate-sized mentor model whose power is concentrated on a specific task to alleviate the issues above.

% 교사가 생성한 데이터는 mutli-step reasoning에 대한 supervison을 제공할 수 있지만, 교사의 낮은 quality on specific task 와 black manner 때문에 insufficient --> 어떤 멘토? : task specific + open sourced intermediate size 로 위 문제를 완화할 것임. 

\paragraph{Training.} For training the mentor model, we directly fine-tune it on the previously constructed $\mathcal{D}_{\text{teacher}}$. Specifically, the mentor model receives $q_i$ as an input, $l^t_i$ as a label, and is trained with a standard language modeling objective.

\paragraph{Rationale Augmentation.} The trained mentor model is then used for train data augmentation. For data samples from $\mathcal{D}$, we let the mentor model annotate step-by-step rationales, given $q_i$ as an input. The mentor in return generates a label $l^m_i$, which consists of a step-by-step rationale and a prediction of its own. We filter the annotations by the mentor identical to filtering the teacher's annotations and preserve data samples where $\hat{y}^m_i = y_i$. Through this stage, we construct $\mathcal{D}_{\text{mentor}}=\{(q_i, l^m_i, y_i)\}_{i=1}^N$ per dataset.

With annotations obtained from the teacher ($\mathcal{D}_{\text{teacher}}$) and the mentor ($\mathcal{D}_{\text{mentor}}$), we finally construct $\mathcal{D}_{\text{train}}$ for training the student model\footnote{It is worth noting that we do not distinguish where the CoT annotations were generated from, but we randomly sample instances from $ \mathcal{D}_{\text{train}}$ to train the student models.}, which is defined as follows:

\begin{equation}
    \mathcal{D}_{\text{train}} = \mathcal{D}_{\text{teacher}} \cup \mathcal{D}_{\text{mentor}}
\end{equation}

\begin{table*}
\centering
\begin{tabular}{l|c|ccc|c}
\toprule
Model & \multicolumn{1}{l}{\#Params} & \multicolumn{1}{|l}{GSM8K} & \multicolumn{1}{l}{ASDiv} & \multicolumn{1}{l|}{SVAMP} & \multicolumn{1}{l}{CommonsenseQA} \\ \midrule
GPT-3.5-Turbo (teacher)* & - & 73.98 & 79.64 & 75.14 & 74.35 \\ \midrule
FlanT5-XXL (mentor) & 11B & 34.34 & 50.32 & 51.71 & 85.01 \\ \midrule
GPT-3-curie \cite{ho-etal-2023-ftcot} & 6.7B & 6.75 & - & 12.67 & 56.76 \\
T5-XXL \cite{magister-etal-2023-teaching} & 11B & 21.99 & 42.12 & - & - \\
FlanT5-XL \cite{fu-etal-2023-specializing} & 3B & 22.40 & 28.40 & 23.80 & - \\
FlanT5-XL (Vanilla-KD)* & 3B & 22.76 & 29.41 & 29.33 & 81.13 \\
FlanT5-XL (MCC-KD)* & 3B & 24.28 & 31.35 & 30.00 & 82.88 \\
\cellcolor[gray]{.9}FlanT5-XL (Mentor-KD (ours)) & \cellcolor[gray]{.9}3B & \cellcolor[gray]{.9}\textbf{24.76} & \cellcolor[gray]{.9}\textbf{31.86} & \cellcolor[gray]{.9}\textbf{32.70} &  \cellcolor[gray]{.9}\textbf{87.14} \\ \bottomrule
\end{tabular}
\caption{Comparison with different baselines on arithmetic and commonsense reasoning tasks. The reported results are averaged accuracy over four runs using randomly selected seeds. Performances marked with an asterisk(*) were excerpted from MCC-KD \cite{chen-etal-2023-mcckd}. The best results are highlighted in \textbf{boldface}.}
\label{tab:main_table}
\end{table*}

\subsection{Reasoning Distillation} \label{sec:reasoning_distillation}
For training the student model, we incorporate both fine-tuning (rationale distillation) and knowledge distillation through logit values obtainable via the mentor model (soft label distillation). This is to allow the student model to jointly 1) learn how to practice step-by-step reasoning in a symbolic manner \cite{ho-etal-2023-ftcot, li-etal-2023-sctod, magister-etal-2023-teaching}, as well as 2) mimic the predictive behavior of a larger model \cite{hinton-etal-2015-kd}. In correspondence, our training objective consists of two loss functions.

\paragraph{Rationale Distillation.}
Identical to training the mentor model, the step-by-step reasoning ability can be distilled through fine-tuning the student model with question-label pairs obtained from the teacher and the mentor. More specifically, the form of learning the multi-step reasoning ability through fine-tuning is defined as follows:
\begin{equation}
    \mathcal{L}_\text{rd} = \mathbbm{E}_{\mathcal{D}_\text{train}}\log{\text{P}_{f}([q; r; y])},
\end{equation}
where $f$ indicates the student model, and the square brackets indicate string concatenation.
%soft label을 통해 얻을 수 있는 지식이 무엇인지가 드러나도록 바꿔야함(reference 걸어서) --> 길게 X
\paragraph{Soft Label Distillation.}
%Access to the teacher model's internal knowledge such as full output distribution is impossible, due to its black-box nature. 
Leveraging the LLM teacher's internal knowledge can be impractical due to its black-box nature or enormous size. 
Instead, we employ our mentor model to provide the soft labels for distillation.
The soft labels are obtained through a forward pass, followed by a softmax function, given $q$ as an input. Formally, we obtain the soft label (probability distribution) $p_k$ of the mentor and student models from the logit value $z_k$ at the $k$-th position through the following equation:
\begin{equation}
    p_k = \frac{\exp{(z_k / \tau)}}{\sum_j \exp({z_j / \tau})},
\end{equation}
where $\tau$ indicates a temperature hyperparameter for softening the distribution.
After obtaining probability distributions of the mentor ($p^m$) and the student ($p^s$), we adopt the Kullback-Leibler divergence loss to minimize the divergence between the two distributions. This allows the student model to mimic the predictive behavior and learn the internal knowledge of larger models. The training objective for soft label distillation is defined as follows:
\begin{equation}
    % \mathcal{L}_\text{kd} = \text{D}_\text{kl}(p_\text{ta} \ || \ p_\text{student})
    \mathcal{L}_\text{sld}(p^m, p^s) =
        \sum_k{p^m_k \log{\frac{p^m_k}{p^s_k}}}
\end{equation}

\paragraph{Joint Learning.}
Finally, we have the student model to jointly learn the aforementioned two objectives. The loss function for training the student model is as follows:
\begin{equation} \label{eq:loss_function}
    \mathcal{L} = (1 - \lambda)\mathcal{L}_\text{rd} + \lambda\mathcal{L}_\text{sld},
\end{equation}
where $\lambda$ is a hyperparameter for interpolating the two loss functions.

\section{Experiments}
In this section, we describe the experiment details and evaluate our Mentor-KD on various complex reasoning tasks.

\subsection{Experiment Setup} \label{sec:experiment_setup}

\paragraph{Tasks and Datasets.} Following \cite{wei-etal-2022-cot, kojima-etal-2022-zscot}, we evaluate our Mentor-KD on four categories of complex reasoning tasks, which are commonsense, arithmetic, logical, and symbolic reasoning. Specifically, we adopt up to three datasets per task in order to evaluate our framework on various datasets of the same task type. Datasets used for this paper are StrategyQA \cite{geva-etal-2021-strategyqa}, CommonsenseQA \cite{talmor-etal-2019-csqa} for commonsense reasoning, GSM8K \cite{cobbe-etal-2021-gsm8k}, ASDiv \cite{miao-etal-2020-asdiv}, and SVAMP \cite{patel-etal-2021-svamp} for arithmetic reasoning, Tracking Shuffled Objects, Date Understanding \cite{srivastava-etal-2023-bigbench} for logical reasoning, and Last Letter Concatenation \cite{wei-etal-2022-cot, kojima-etal-2022-zscot} for symbolic reasoning. Further details are provided in Appendix~\ref{sec:dataset_stats}.

\paragraph{Language Models.} We utilize \texttt{gpt-3.5-turbo} through OpenAI API for our teacher model. 
For the mentor and student models, we mainly use FlanT5-XXL and FlanT5-XL \cite{chung-etal-2022-flant5} as our mentor and student models. For additional analysis, we use various sizes of FlanT5 and T5 \cite{raffel-etal-2020-t5}, including large, base, and small-sized models.

%and incorporate T5 \cite{raffel-etal-2020-t5} to our additional experiments. Specifically, we adopt FlanT5-XXL and FlanT5-XL as our mentor and student models for our main experiments, respectively.

\paragraph{Chain-of-Thought Annotations.} For GSM8K, ASDiv, SVAMP, and CommonsenseQA, we utilize the CoT annotations provided by \cite{chen-etal-2023-mcckd}. The annotations were collected with GPT-3.5-Turbo using Zero-shot-CoT prompting, which is identical to our methodology mentioned in Section~\ref{sec:cot_annotation}. Other datasets were newly prompted and collected by our research institute.

\begin{table*}[t]
\centering
\resizebox{\textwidth}{!}{%
\begin{tabular}{lclccccccc}
\hline
\multirow{2}{*}{Model} & \multicolumn{1}{c}{\multirow{2}{*}{\#Params}} & \multirow{2}{*}{Method} & \multicolumn{2}{c}{Commonsense} & \multicolumn{2}{c}{Arithmetic} & \multicolumn{2}{c}{Logical} & Symbolic \\ \cline{4-10} 
 & \multicolumn{1}{c}{} &  & SQA & CSQA & ASDiv & SVAMP & Shuffled & Date & Last Letter \\ \hline
GPT-3.5-Turbo & - & ZS-CoT (teacher) & 58.07 & 74.35* & 79.64* & 75.14* & 64.00 & 81.98 & 68.00 \\ \hline
T5-large & 780M & Vanilla-KD (mentor) & 63.32 & 68.80 & 12.42 & 13.05 & 90.22 & 84.68 & 68.00 \\ \hline
\multirow{3}{*}{T5-base} & \multirow{3}{*}{250M} & Vanilla-KD & 61.43 & 55.53 & \underline{11.15} & \textbf{10.00} & \underline{77.33} & \textbf{89.19} & \underline{56.00} \\
 &  & MCC-KD & \underline{62.01} & \underline{57.17} & 9.55 & \underline{8.00} & 56.89 & 81.98 & 45.33 \\
 &  & \cellcolor[gray]{.9}Mentor-KD (ours) & \cellcolor[gray]{.9}\textbf{62.45} & \cellcolor[gray]{.9}\textbf{59.05} & \cellcolor[gray]{.9}\textbf{12.10} & \cellcolor[gray]{.9}\textbf{10.00} & \cellcolor[gray]{.9}\textbf{92.00} & \cellcolor[gray]{.9}\underline{88.29} & \cellcolor[gray]{.9}\textbf{65.33} \\ \hline
\multirow{3}{*}{T5-small} & \multirow{3}{*}{80M} & Vanilla-KD & 55.60 & \underline{42.75} & 5.10 & 6.67 & \underline{39.11} & \underline{81.98} & \underline{48.67} \\
 &  & MCC-KD & \underline{56.77} & 38.25 & \underline{5.73} & \underline{7.33} & 38.22 & 77.48 & 28.67 \\
 &  & \cellcolor[gray]{.9}Mentor-KD (ours) & \cellcolor[gray]{.9}\textbf{57.93} & \cellcolor[gray]{.9}\textbf{45.37} & \cellcolor[gray]{.9}\textbf{7.01} & \cellcolor[gray]{.9}\textbf{8.67} & \cellcolor[gray]{.9}\textbf{79.56} & \cellcolor[gray]{.9}\textbf{87.39} & \cellcolor[gray]{.9}\textbf{56.67} \\ \hline
\end{tabular}%
}
\caption{Performances of teacher, mentor, and student models across four different complex reasoning tasks, where the backbone model is T5. GPT-3.5-Turbo results with an asterisk(*) were excerpted from \cite{chen-etal-2023-mcckd}. The best and second best results are highlighted in \textbf{boldface} and \underline{underline}, respectively.}
\label{tab:slms_t5}
\end{table*}

\begin{table*}[t]
\centering
\resizebox{\textwidth}{!}{%
\begin{tabular}{lclccccccc}
\hline
\multirow{2}{*}{Model} & \multicolumn{1}{c}{\multirow{2}{*}{\#Params}} & \multirow{2}{*}{Method} & \multicolumn{2}{c}{Commonsense} & \multicolumn{2}{c}{Arithmetic} & \multicolumn{2}{c}{Logical} & Symbolic \\ \cline{4-10} 
 & \multicolumn{1}{c}{} &  & SQA & CSQA & ASDiv & SVAMP & Shuffled & Date & Last Letter \\ \hline
GPT-3.5-Turbo & - & ZS-CoT (teacher)& 58.07 & 74.35* & 79.64* & 75.14* & 64.00 & 81.98 & 68.00 \\ \hline
FlanT5-large & 780M & Vanilla-KD (mentor)& 64.48 & 79.36 & 20.70 & 14.00 & 90.22 & 88.29 & 65.33 \\ \hline
\multirow{3}{*}{FlanT5-base} & \multirow{3}{*}{250M} & Vanilla-KD & 62.74 & 62.33 & 12.42 & 10.67 & \underline{84.89} & \underline{86.49} & \underline{53.33} \\
 &  & MCC-KD & \underline{64.92} & \textbf{68.47} & \underline{13.69} & \textbf{12.00} & 69.78 & 85.59 & 46.00 \\
 &  & \cellcolor[gray]{.9}Mentor-KD (ours) & \cellcolor[gray]{.9}\textbf{65.21} & \cellcolor[gray]{.9}\underline{67.24} & \cellcolor[gray]{.9}\textbf{15.29} & \cellcolor[gray]{.9}\underline{11.33} & \cellcolor[gray]{.9}\textbf{93.78} & \cellcolor[gray]{.9}\textbf{87.39} & \cellcolor[gray]{.9}\textbf{65.33} \\ \hline
\multirow{3}{*}{FlanT5-small} & \multirow{3}{*}{80M} & Vanilla-KD & 55.90 & \underline{48.24} & \underline{7.96} & \textbf{10.67} & \underline{63.11} & \textbf{85.59} & \underline{52.67} \\
 &  & MCC-KD & \underline{58.37} & 45.21 & 7.01 & \underline{10.00} & 43.11 & 81.98 & 35.33 \\
 &  & \cellcolor[gray]{.9}Mentor-KD (ours) & \cellcolor[gray]{.9}\textbf{59.97} & \cellcolor[gray]{.9}\textbf{48.98} & \cellcolor[gray]{.9}\textbf{10.83} & \cellcolor[gray]{.9}\textbf{10.67} & \cellcolor[gray]{.9}\textbf{82.67} & \cellcolor[gray]{.9}\underline{83.78} & \cellcolor[gray]{.9}\textbf{58.67} \\ \hline
\end{tabular}%
}
\caption{Performances of teacher, mentor, and student models across four different complex reasoning tasks, where the backbone model is FlanT5. GPT-3.5-Turbo results with an asterisk(*) were excerpted from \cite{chen-etal-2023-mcckd}. The best and second best results are highlighted in \textbf{boldface} and \underline{underline}, respectively.}
\label{tab:slms_flant5}
\end{table*}

\begin{table}[t]
\centering
\resizebox{\columnwidth}{!}{
\begin{tabular}{clcc}
\hline
Model & \multicolumn{1}{c}{Method} & Shuffled & Last Letter \\ \hline
\multirow{3}{*}{T5} &\cellcolor[gray]{.9} Mentor-KD (ours) & \cellcolor[gray]{.9}\textbf{79.56} & \cellcolor[gray]{.9}\textbf{56.67} \\
 & \;\;w/o RD & 32.89 & 50.00 \\
 & \;\;w/o SLD & 76.00 & 52.00 \\ \hline
\multirow{3}{*}{FlanT5} & \cellcolor[gray]{.9} Mentor-KD (ours) & \cellcolor[gray]{.9}\textbf{82.67} & \cellcolor[gray]{.9}\textbf{58.67} \\
 & \;\;w/o RD & 64.89 & 56.00 \\
 & \;\;w/o SLD & 82.22 & 54.00 \\ \hline
\end{tabular}
}
\caption{Ablation study of Mentor-KD on Tracking Shuffled Objects and Last Letter Concatenation. We employ \texttt{large} models of each backbone model as mentors and \texttt{small} models as students.}
\label{tab:ablation}
\end{table}

% KARD 참고
\paragraph{Baselines.} For the baselines, we incorporate previous methods of reasoning distillation. Specifically, we implement Vanilla-KD, a general reasoning distillation method that fine-tunes student models on the teacher model's generated rationales \cite{ho-etal-2023-ftcot, magister-etal-2023-teaching}, and MCC-KD, which further emphasizes diversity and consistency within multiple CoT rationales \cite{chen-etal-2023-mcckd}. We also compare Mentor-KD's performances with \citet{fu-etal-2023-specializing}, which aims to specialize LM's reasoning ability towards a specific task. We report the teacher model's performances via Zero-shot-CoT (ZS-CoT) prompting.
%The baselines of the above show the capability of student models to practice multi-step reasoning by utilizing distillation sets from the LLM alone. 

 % To evaluate the effectiveness of Mentor-KD, we provide a comparison of our method with various existing reasoning distillation methods which leverage LLM as teacher models to generate rationales and distill their reasoning abilities into student models. Specifically, we mainly implement Vanilla-KD, a method to fine-tune student models on the teacher model's generated rationales \cite{ho-etal-2023-ftcot, magister-etal-2023-teaching}, and MCC-KD, which further emphasizes diversity and consistency during reasoning distillation \cite{chen-etal-2023-mcckd}. We also compare Mentor-KD's performances with \citet{fu-etal-2023-specializing} which aims to specialize LM's reasoning ability towards a specific task.

\paragraph{Implementations.} We adopt models provided by HuggingFace \cite{wolf-etal-2020-huggingface} on two NVIDIA RTX A6000 GPUs. Specifically, we train models for 18 epochs for XXL-/XL-sized models, 10 epochs for large, and 20 epochs for base, and small models following the previous works \cite{chen-etal-2023-mcckd, ho-etal-2023-ftcot}. The maximum sequence length is set to 512 throughout all our experiments, and we sweep batch sizes in \{2, 4, 6, 8\}. To accelerate training and conserve memory usage, we apply mixed precision of \texttt{bfloat16} and LoRA \cite{lora} throughout our main experiments and follow the related configurations from  \cite{chen-etal-2023-mcckd}. Moreover, We use AdamW \cite{loshchilov-2019-adamw} optimizer, with a learning rate of \{1e-4, 2e-4, 3e-4, 5e-4\}. We apply the loss interpolation hyperparameter $\lambda$ to 0.3, and the distillation temperature $\tau$ to \{1.0, 2.0\}. We report the average test accuracy results from four random seeds.

\subsection{Main Results}
For a fair comparison, we mainly compare Mentor-KD utilizing FlanT5-XL models on three arithmetic reasoning tasks and one commonsense reasoning task, which are commonly used in reasoning distillation \cite{ho-etal-2023-ftcot, chen-etal-2023-mcckd}. % In addition, we use the training and test datasets from MCC-KD \cite{chen-etal-2023-mcckd}.
%For a fair comparison, we evaluate the performance of our student models by utilizing the training and test datasets from MCC-KD \cite{chen-etal-2023-mcckd}.
%each test dataset from MCC-KD \cite{chen-etal-2023-mcckd}.
The main results are provided in Table \ref{tab:main_table}. We observe that our Mentor-KD achieves state-of-the-art performance on four different reasoning datasets. Specifically, our model achieves approximately 2.0\% better performance on averaged accuracy than MCC-KD, the previous SOTA model. The results demonstrate the effectiveness of Mentor-KD in addressing challenging complex reasoning tasks, including both arithmetic and commonsense reasoning.

\section{Analysis}
To delve into the benefits of our method, we perform a series of fine-grained analytical experiments with the following research questions (RQs):
\setlist{nolistsep}
\begin{itemize}[leftmargin=*, noitemsep]
\item \textbf{RQ1.} Can Mentor-KD be generalized to the various sizes and types of student models? (\S \ref{sec:5.1})
\item \textbf{RQ2.} How does each component in Mentor-KD contribute to its overall performance? (\S \ref{sec:5.2})
%\item \textbf{RQ2.} Is CleaR capable of effectively learning from clean samples while minimizing the influence of noisy ones? (\S 5.2)
\item \textbf{RQ3.} Can the mentor model generate informative distillation sets for students? (\S \ref{sec:5.3})
\item \textbf{RQ4.} Does Mentor-KD offer improvements under low-resource scenarios? (\S \ref{sec:5.4})
\item \textbf{RQ5.} Does the size of mentor models affect the performance of student models? (\S \ref{sec:5.5})
\end{itemize}

\begin{figure}[t]
\centering
    \includegraphics[width=0.48\textwidth]{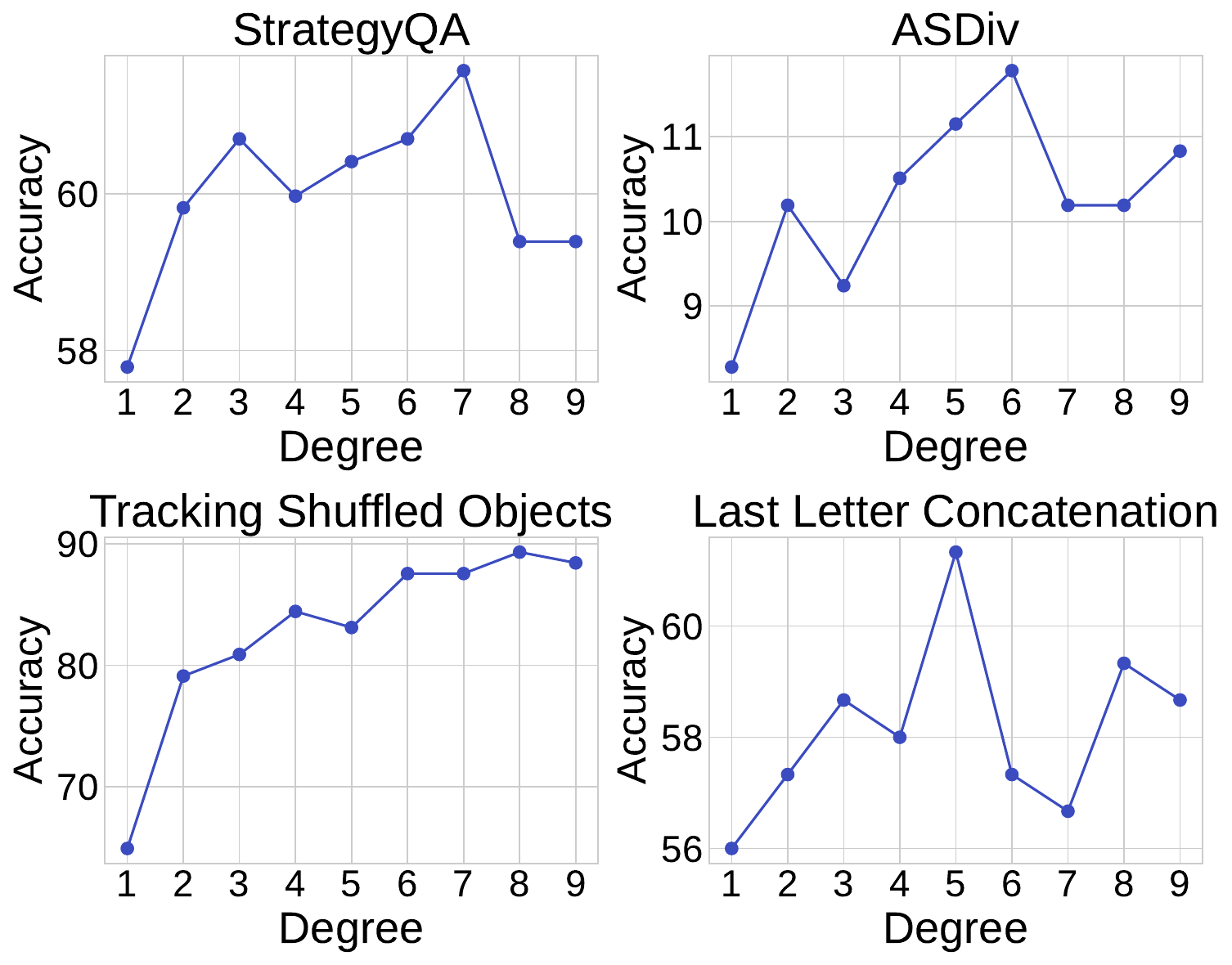}
\caption{Performances by differentiating the degree (number) of mentor-generated CoT rationales per question. We adopt FlanT5-large and FlanT5-small as mentor and student models, respectively.}
\label{fig:degree_of_augmentation}
\end{figure}

\subsection{Various Student Models (RQ1)}
\label{sec:5.1}
To further investigate the generality of our Mentor-KD, we conduct experiments on various types of student models with different sizes. Notably, we further expand our scope of experiments by additionally incorporating logical and symbolic reasoning tasks. Specifically, we utilize T5 and FlanT5, which are widely adopted in LLM distillation following previous works \cite{ho-etal-2023-ftcot, chen-etal-2023-mcckd}. We leverage \texttt{large} variants of T5 and FlanT5 as our mentor model, and \{\texttt{base}, \texttt{small}\} variants as our student model. Details on implementations of this section are elaborated in Appendix~\ref{sec:slms}.

The results are shown in Tables \ref{tab:slms_t5} and \ref{tab:slms_flant5}. We observe that our Mentor-KD consistently outperforms the other baselines in four categories of complex reasoning tasks on various student models. In particular, Mentor-KD has shown large performance improvements in commonsense and logical reasoning tasks, which the student model may even outperform the performances of the LLM teacher (i.e., GPT-3.5). These results demonstrate that our task-specific mentor model can successfully complement the insufficient LLM teacher's knowledge, thereby leading to achieving better performances for various student models by transferring more informative distillation signals.
%performance to student models.
%model successfully mitigate the distillation data deficiency issue from the imperfect LLM teachers.

\begin{figure}[t]
\centering
    \includegraphics[width=0.48\textwidth]{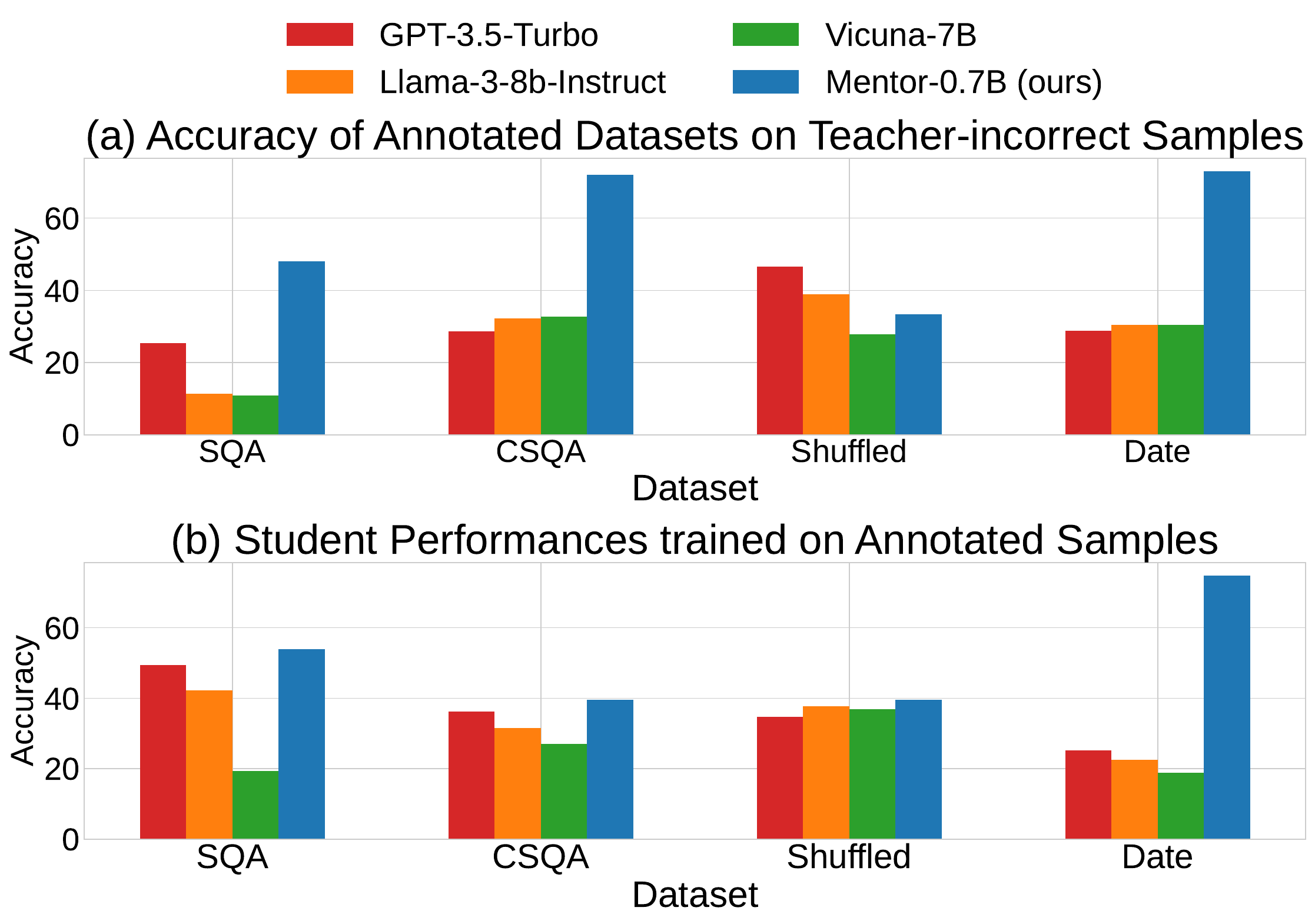}
\caption{Comparison of (a) accuracy of our mentor model (FlanT5-large) and LLM baselines on teacher-incorrect samples, and (b) performances of student models trained with augmented distillation sets from LLM baselines and our mentor models.}
\label{fig:aug_stats}
\end{figure}

\subsection{Ablation Studies (RQ2)}
\label{sec:5.2}
We conduct ablation studies to explore the contributions brought by each technique of our method. Specifically, we focus on the effect of rationale distillation (RD) and soft label distillation (SLD) from the mentor model. The detailed results are shown in Table ~\ref{tab:ablation}. We observe that omitting RD and SLD significantly affects both model types and datasets. These results emphasize the significance of RD for both training samples and soft labels, which enhance the insufficient knowledge from teachers.
% 두가지의 다른 종류의 교사 지식이 모두 중요하다는 방향으로 작성

%The results demonstrate that each component is essential for improving the reasoning ability. 

%ablate one component out of the two roles that the mentor model practices (i.e., data augmentation and soft label provision).

%Table~\ref{tab:ablation} represents the results. We initially observe that Mentor-KD's performances are preserved when both components are taken into consideration, which highlights the significance of both components. We further observe that the performance loss is greater when data augmentation is ablated in general. This indicates that of the two roles that the mentor model practices, data augmentation contributes more to Mentor-KD.

\subsection{Impact of Data Augmentation (RQ3)}
To further investigate the proposed data augmentation methods of mentor models, we additionally analyze the effectiveness in perspectives of both quantity and quality.
%We investigate the effectiveness of data augmen Mentor-KD with 
\label{sec:5.3}
\paragraph{Quantity of Augmented Dataset.} We first analyze the impact of the number of generated distillation sets from the mentor by diversifying the number of rationales that the mentor produces per question. The results are shown in Figure \ref{fig:degree_of_augmentation}. Generally, we observe that student performances improve in line with the quantity of distillation sets. This indicates that our mentor models successfully generate rationales helpful for student models to learn multi-step reasoning. However, we also observe the performance usually saturated over six augmentations and begins to decline when more distillation sets are introduced, which may be due to the noises generated from models \cite{gkp}. 

%while they are saturated over five augmentations. 
%Then, the performance declines when more distillation sets are introduced, which may be because repetitive training samples are generated by mentor models. Nevertheless, the results demonstrate the augmented distillation sets are effective for the student models.

\paragraph{Quality of Augmented Dataset.} To investigate the quality of our augmented distillation sets, we compare our mentor models (i.e., FlanT5-large) with various LLMs that may be potential alternatives of mentors for augmentation (i.e. GPT-3.5-Turbo\footnote{We adopt a different seed value from the initial CoT annotation phase (Section \ref{sec:cot_annotation}) for this experiment.}, Llama-3-8B-Instruct\footnote{\url{https://ai.meta.com/blog/meta-llama-3/}}, and Vicuna-7B \cite{chiang-etal-2023-vicuna}). 
We first compare the accuracy of the augmentations mentors generate with other baselines (through Zero-shot-CoT prompting) on incorrect samples predicted by the LLM teacher. We then report the performances of the student (i.e., FlanT5-small) trained on each augmentation to analyze whether task-specific mentors can provide informative sets to the students.

% the teacher model (i.e., InstructGPT-175B), while also incorporating other LMs that may be potential alternatives of mentors for augmentation (i.e., Llama-3-8B-Instruct\footnote{\url{https://ai.meta.com/blog/meta-llama-3/}}, and Vicuna-7B \cite{chiang-etal-2023-vicuna}).

% Since better-performing models can generate more useful distillation sets, we compare the accuracy of the mentor with other baselines on incorrect samples predicted by the LLM teacher. We also report the performance of student models which are trained with the augmented datasets from the mentor and each baselines as well.

% To investigate the quality of our augmented distillation sets, we further compare our mentor models (i.e., FlanT5-large) with various-sized LLMs, including Llama-3-8B-Instruct\footnote{\url{https://ai.meta.com/blog/meta-llama-3/}}, Vicuna-7B \cite{chiang-etal-2023-vicuna}, and InstructGPT-175B, which can also generate additional distillation sets for student models. Since more accurate models can generate more useful distillation sets, we compare the accuracy of the mentor with baselines on incorrect samples predicted by the LLM teacher. We also report the performance of student models, which are trained with the augmented datasets from each baseline and mentor.
%Viccuna, LLaMA, GPT API

The results are shown in Figure \ref{fig:aug_stats}. While the mentor models consist of smaller parameters than the LLMs (e.g., $10\times$ smaller than Llama3-8B-Instruct), they generate more accurate rationales than other LLM baselines, indicating the ability to provide more diverse rationales for student models. 
%significantly outperform LLM baselines on teacher-incorrect samples. 
In addition, we observe that the students trained with distillation sets from mentor models indeed achieve higher performance than those trained with sets from LLM teachers. These results suggest that mentors can generate higher-quality rationales than LLM teachers. Overall results highlight the superiority of task-specific fine-tuning of mentor models.

\begin{figure}[t]
\centering
    \includegraphics[width=0.48\textwidth]{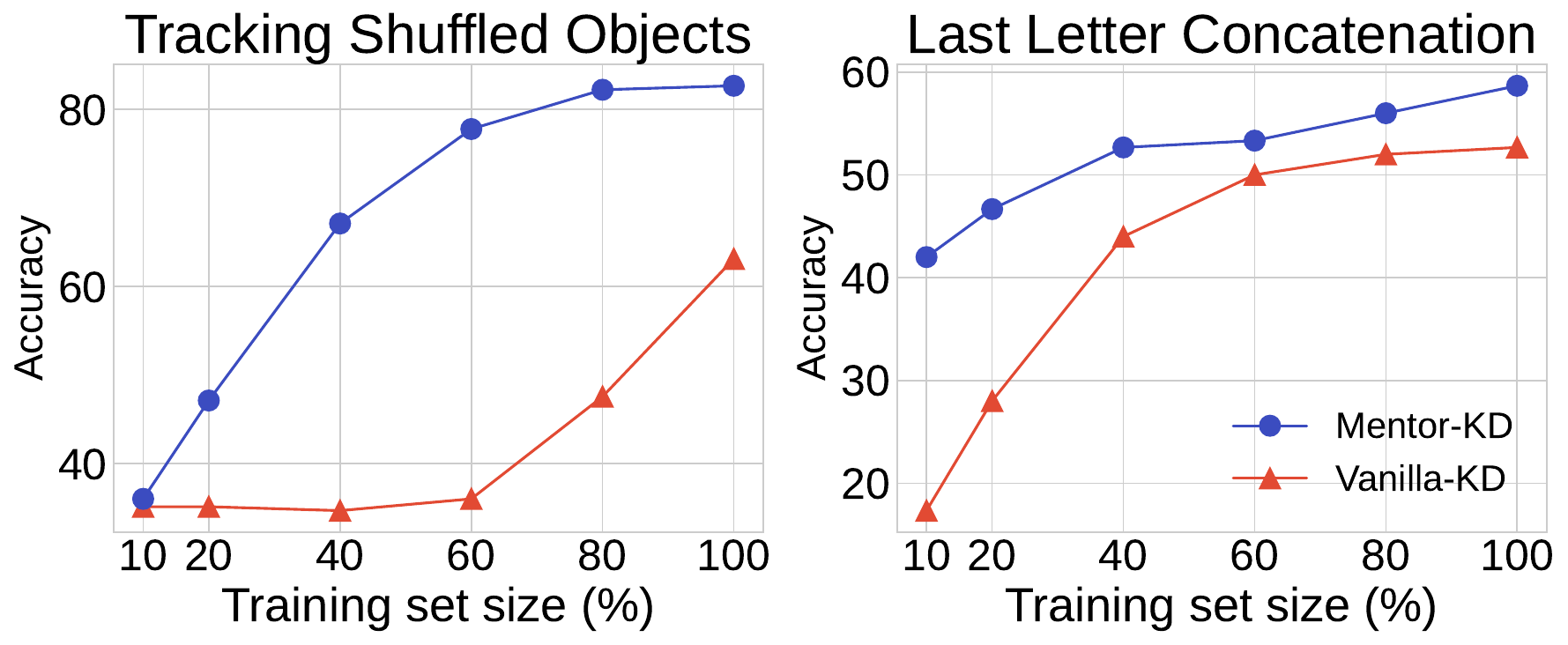}
\caption{Comparison between Mentor-KD (Ours) and Vanilla-KD baseline on various distillation sets by differentiating the percentage of rationales being used.}
\label{fig:partial_data}
\end{figure}

\subsection{Low-resource Scenarios (RQ4)}
\label{sec:5.4}
In reasoning distillation, collecting sufficiently large distillation sets can be prohibitively expensive due to the cost of API calls for black-box LLMs. Therefore, we examine the effectiveness of Mentor-KD on low-resource scenarios, where distillation sets are collected for only a proportion of the original datasets. Specifically, we compare the Vanilla-KD baseline with our Mentor-KD, varying the ratio of distillation sets generated from LLM teacher models. The results are shown in Figure \ref{fig:partial_data}.

%The results of two reasoning tasks are shown in Figure \ref{fig:partial_data}.

% In LLM distillation, collecting sufficiently large distillation sets can be prohibitively expensive, due to the cost of API calls for black-box LLMs. We therefore examine the effectiveness of Mentor-KD on low-resource scenarios, which have insufficient distillation sets. Specifically, we compare the fine-tuned CoT baseline and our model varying the ratio of distillation sets generated from LLM teacher models. The results of four reasoning tasks are shown in Figure 3.

We observe that the Mentor-KD also allows performance improvements for student models in low-resource scenarios, given that mentor models provide informative rationale sets and soft labels. In particular, the Vanilla-KD baseline shows performance degradation on highly limited distillation signals, while our Mentor-KD exhibits robustness for limited datasets. These results demonstrate that our mentor models can alleviate over-fitting problems for students from the limited distillation signals and can distill the LLM teacher's knowledge in a cost-efficient manner. We elaborate on this research question in Appendix \ref{sec:additional_costs}.
%These results demonstrate that our mentor models can alleviate the over-fitting for limited distillation signals and distill the LLM teacher's knowledge in a manner of cost-efficiency.
%These results also demonstrate our mentor models alleviate the that over-fitted teachers adversely affect the performance of the student model and that PEFT can either be a promising method to alleviate this with its robustness on over-fitting.

\begin{figure}[t]
\centering
    \includegraphics[width=0.48\textwidth]{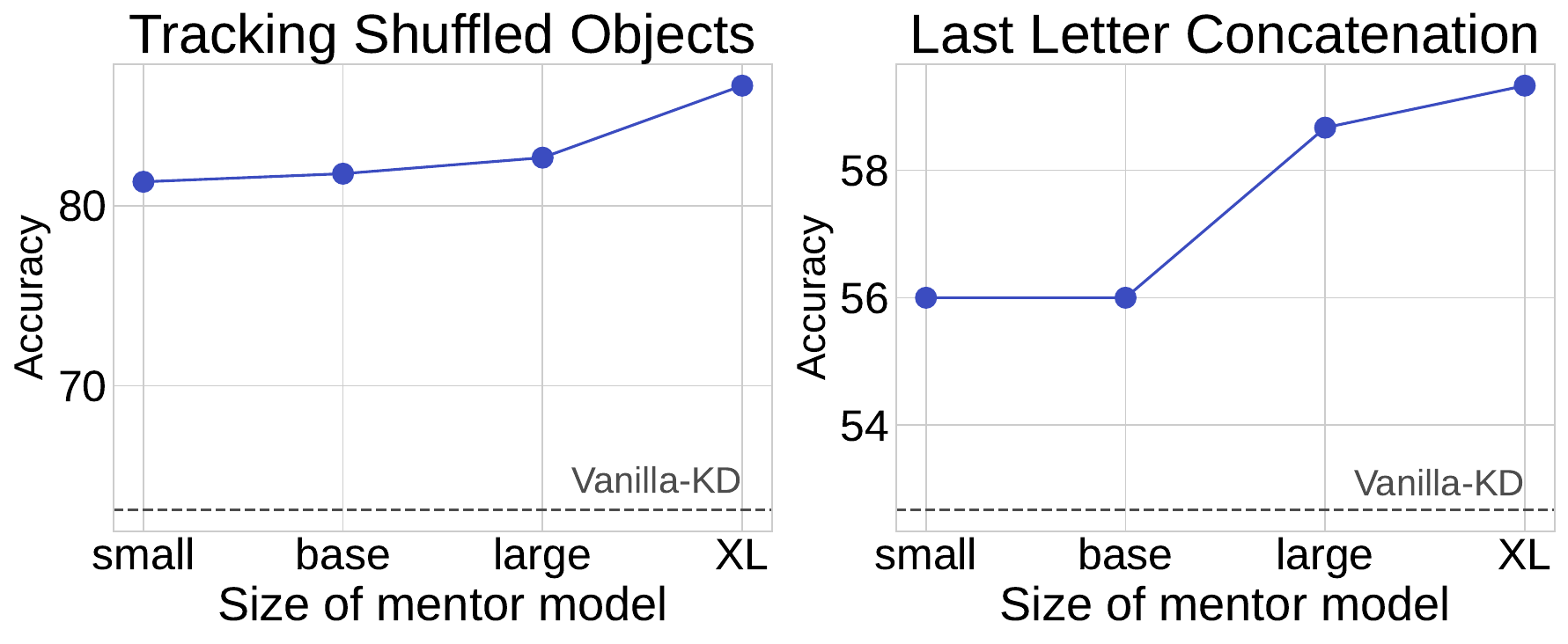}
\caption{Comparison between student (FlanT5-small) performance using different mentor models considering various capacity gap sizes. Dotted lines in gray indicate Vanilla-KD baseline performances.}
\label{fig:design_choice}
\end{figure}

\subsection{Effects of Mentor Sizes (RQ5)}
\label{sec:5.5}
% 모델 크기에 따른 성능 분석
% 지식증류에서는 더 큰 교사가 꼭 더 좋은 성능을 내는 것이 아니며, capacity gap 차이가 클수록 성능이 하락하는 경향이 있다. 따라서, 우리는 최적의 교사 모델에 대한 해석을 제공하기 위해 다양한 사이즈의 mentor 모델에 대한 실험을 수행한다. 그 결과는 ...

To further explore Mentor-KD's effectiveness and verify our design choice, we conduct an additional experiment by differentiating the size of mentor models. Here, we employ FlanT5-small as a student model and FlanT5-\{XL, large, base, small\} as mentor models. For distilling \texttt{small} to \texttt{small} models, we utilize self-distillation, following previous works \cite{allen-zhu-etal-2023-self_distillation, zhu-etal-2024-pad}.

Figure \ref{fig:design_choice} displays the results. Generally, we observe that the student model performs better when larger mentor models are incorporated during reasoning distillation. Employing the smallest mentor results in a performance decline, but we observe such scenarios still outperform the baselines in Table \ref{tab:slms_flant5}. The results suggest that employing larger models of better performances contributes to boosting the small student models' performances, which is aligned with previous findings that student performances are correlated to their corresponding model's performances \cite{ho-etal-2023-ftcot}.

% Conversely, the performances tend to decline when adopting smaller mentor models. Nevertheless, it is worth noting that despite the performance decline, we still observe scenarios that employ the smallest mentors outperform the baselines in Table \ref{tab:slms_flant5}. The results suggest that employing larger models of better performances contributes to boosting the small student models' performances, which is aligned with previous findings that student performances are correlated to their corresponding model's performances \cite{ho-etal-2023-ftcot}.

\section{Conclusion}
We have presented Mentor-KD, a novel framework to transfer reasoning capabilities from LLMs to smaller LMs.
%by complementing the limited applicability for training datasets of LLMs. 
To this end, we have introduced a mentor model, a novel auxiliary model, for complementing the distillation sets from LLMs by augmenting multi-step rationales and providing soft labels for the student model. Through extensive experiments, we have demonstrated that our Mentor-KD significantly improves the effectiveness of reasoning distillation. Specifically, our student models outperform existing reasoning distillation baselines with various sizes and types of models on complex reasoning tasks. Furthermore, we have verified that our mentor model can generate effective reasoning samples and soft labels for training student models, resulting in consistent performance improvements.

\section{Limitations}
% In this section, we discuss the limitations of our work which can be summarized into two points. First, among a variety of formats for multi-step reasoning, we mainly focus on those in a natural language format. However, recent studies have begun paving ways to distil LLM's step-by-step reasoning capabilities through reasoning programs \cite{zhu-etal-2024-pad}, and it is an open question whether Mentor-KD can improve on such strategies. We leave this question as a future research direction.

% Next, the evaluation scheme on the correctness of generated CoT annotations is primarily based on string match between final predictions of the model and the golden labels, which many works adopt \cite{ho-etal-2023-ftcot, li-etal-2023-sctod, wang-etal-2023-democratizing, fu-etal-2023-specializing, shridhar-etal-2023-distilling}. Although it is reported that such criterion allows the alignment between the correctness of rationales and answers \cite{chen-etal-2023-mcckd}, we posit that discussions regarding the trustworthiness of this criterion should be dealt with more in-depth in this field.

% Cost, Model, Further Applications
While we have demonstrated that Mentor-KD effectively improves the reasoning ability of small language models by augmenting both training sets and soft labels, there are some limitations that present promising avenues for future research. 

\paragraph{Training Costs for Mentor Models.} Our framework requires additional computational costs for training mentor models for reasoning distillation. Besides the training costs in the distillation process, this study mainly focuses on improving the inference efficiency of small student models, as with most reasoning distillation research \cite{ho-etal-2023-ftcot, chen-etal-2023-mcckd, wang-etal-2023-scott}.  We further elaborate on this issue in Appendix~\ref{sec:additional_costs}.

\paragraph{Exploration on Different Reasoning Strategies.} While we successfully demonstrate the performance improvements in CoT reasoning abilities for small language models, it is an open question whether our framework can be applied to other types of reasoning strategies, such as program-guided reasoning \cite{zhu-etal-2024-pad}, retrieval-based reasoning \cite{kang-etal-2023-kard, zhao-etal-2024-prr}, and reasoning based on contextualized, structured knowledge \cite{park-etal-2024-coconut}. We leave the exploration of distillation for various types of reasoning strategies as a future research direction in this field.

\paragraph{Exploration on Different Architectures.} We have verified the effectiveness of our framework on encoder-decoder models (e.g., FlanT5, T5) with fewer than 3 billion parameters as the student models. Therefore, the applicability of our framework to decoder-only models remains under-explored in this work. Nevertheless, based on recent evidence suggesting that reasoning distillation can be effectively generalized to various architectures \cite{ho-etal-2023-ftcot, chen-etal-2023-mcckd, wang-etal-2023-democratizing}, we believe that Mentor-KD is expected to display performance boosts on decoder-based student models as well.

\section*{Acknowledgements}
This work was supported by the National Research Foundation of Korea (NRF) grant funded by the Korea government (MSIT) (No.RS-2024-00415812 and No.2021R1A2C3010430) and Institute of Information \& communications Technology Planning \& Evaluation (IITP) grant funded by the Korea government (MSIT) (No.RS-2024-00439328, Karma: Towards Knowledge Augmentation for Complex Reasoning (SW Starlab), No.RS-2024-00457882, AI Research Hub Project, and No.RS-2019-II190079, Artificial Intelligence Graduate School Program (Korea University)).

% \bibliography{main}

\clearpage
\newpage
\newpage
\appendix

\begin{center}
\LARGE
\textbf{Appendix}    
\end{center}
\label{sec:appendix}

\section{Dataset Statistics} \label{sec:dataset_stats}

We provide the statistics of the datasets implemented in our study in Table~\ref{tab:dataset_stats}, including their original licenses. We follow the train-test dataset splits for GSM8K, ASDiv, SVAMP, and CommonsenseQA from \cite{chen-etal-2023-mcckd}. For StrategyQA, Tracking Shuffled Objects, Date Understanding, and Last Letter Concatenation, we follow the train-test dataset splits from \cite{ho-etal-2023-ftcot}.

Meanwhile, in practice, we utilize CoT annotations from \cite{chen-etal-2023-mcckd} for GSM8K, ASDiv, SVAMP, CommonsenseQA, and newly prompt the LLM for other datasets. For other datasets, we prompt the LLM of six CoT annotations per question. Furthermore, we report in Table~\ref{tab:aug_stats} the number of CoT rationales augmented (the size of $\mathcal{D}_\text{mentor}$) by our mentor model (FlanT5-large) that has been used in experiments of Section~\ref{sec:5.1}.

\section{Implementation Details on Various Student Models} \label{sec:slms}
For experiments on models smaller than 1B, we use T5 and FlanT5 as our backbone models with an AdamW optimizer. We conduct a hyperparameter search on $\tau$ of \{1.0, 1.5, 2.0\}, $\lambda$ of \{0.1, 0.2, 0.3, 0.4\}, and learning rate of \{1e-4, 2e-4, 3e-4, 4e-4, 5e-4\}, and report the best test accuracy per epoch. Meanwhile, for labels of the question-label pairs, we adopt the template ``\{$r_i$\}. $--$> \{$y_i$\}.'' for saving tokenization spaces following \cite{ho-etal-2023-ftcot}. For experiments on the Vanilla-KD baseline and our Mentor-KD, we randomly select three out of six CoT annotations per question. Moreover, we have the mentor model generate three CoT rationales per question for augmentation.

\section{Additional Costs for Mentor Models} \label{sec:additional_costs}

\begin{table}[h]
\centering
\begin{tabular}{cccc}
\hline
Method & Train Set & Shuffled & Last Letter \\ \hline
Vanilla-KD & 100\% & 63.11 & 52.67 \\ \hline
\multirow{3}{*} {\begin{tabular}[c]{@{}c@{}}Mentor-KD \\ (ours)\end{tabular}}  & 100\% & 82.67 & 58.67 \\
 & 80\% & 82.22 & 56.00 \\
 & 40\% & 67.11 & 52.67 \\ \hline
\end{tabular}
\caption{Comparison between Vanilla-KD and Mentor-KD (ours) with different training set ratios.}
\label{tab:partial_data}
\end{table}

Although our study mainly spotlights the inference efficiency of small LMs as mentioned in the limitations section, it may be argued that Mentor-KD requires extra computational costs for training the mentor models. 

However, considering that Mentor-KD achieves comparable performance with smaller distillation sets from LLM teachers, we suggest that Mentor-KD might be more efficient for training the student models than the baselines. This is especially significant, in regard to the substantial inference cost of LLMs (teacher models) \cite{ding-etal-2024-hybrid, wan-etal-2024-efficient}. Specifically, Table~\ref{tab:partial_data} shows that Mentor-KD works on par, or even exceeds the Vanilla-KD baseline trained on 100\% of the distillation sets from the LLM teacher while utilizing only 40\% of them (More detailed results are shown in Figure ~\ref{fig:partial_data}). This indicates the potential to save the inference cost of generating 60\% of the distillation sets by the LLM teacher. Taking the entire KD pipeline into account, Mentor-KD may train the student more efficiently depending on the design choices, such as the size of the mentor models and the number of distillation sets from the LLM.

\begin{table*}
\centering
\resizebox{\textwidth}{!}{%
\begin{tabular}{lcrrcl}
\hline
\multicolumn{1}{c}{Dataset} & Choices & \multicolumn{1}{c}{\# Train Data} & \multicolumn{1}{c}{\# Test Data} & License & \multicolumn{1}{c}{References} \\ \hline
StrategyQA & 2 & 1603 & 687 & Apache-2.0 & \citeauthor{geva-etal-2021-strategyqa} \citeyear{geva-etal-2021-strategyqa}  \\
CommonsenseQA & 5 & 8520 & 1221 & Unspecified & \citeauthor{talmor-etal-2019-csqa} \citeyear{talmor-etal-2019-csqa}  \\
ASDiv & - & 1462 & 314 & CC BY-NC 4.0 & \citeauthor{miao-etal-2020-asdiv} \citeyear{miao-etal-2020-asdiv}  \\
SVAMP & - & 700 & 150 & MIT & \citeauthor{patel-etal-2021-svamp} \citeyear{patel-etal-2021-svamp}  \\
GSM8K & - & 7473 & 659 & MIT & \citeauthor{cobbe-etal-2021-gsm8k}, \citeyear{cobbe-etal-2021-gsm8k}  \\
Tracking Shuffled Objects & 3 & 525 & 225 & Apache-2.0 & \citeauthor{srivastava-etal-2023-bigbench} \citeyear{srivastava-etal-2023-bigbench}  \\
Date Understanding & 5-6 & 258 & 111 & Apache-2.0 & \citeauthor{srivastava-etal-2023-bigbench} \citeyear{srivastava-etal-2023-bigbench}  \\
Last Letter Concatenation & - & 350 & 150 & Unspecified & \citeauthor{wei-etal-2022-cot} \citeyear{wei-etal-2022-cot}; \citeauthor{kojima-etal-2022-zscot} \citeyear{kojima-etal-2022-zscot}  \\ \hline
\end{tabular}%
}
\caption{Statistics of datasets used in our study.}
\label{tab:dataset_stats}
\end{table*}

\begin{table}
\centering
\resizebox{\columnwidth}{!}{%
\begin{tabular}{lrr}
\hline
\multicolumn{1}{c}{Dataset} & \multicolumn{1}{c}{\# Train Data} & \multicolumn{1}{c}{\# Aug Data} \\ \hline
StrategyQA & 1603 & 4396 \\
CommonsenseQA & 8520 & 25413 \\
ASDiv & 1462 & 2667 \\
SVAMP & 700 & 1558 \\
Tracking Shuffled Objects & 525 & 1392 \\
Date Understanding & 258 & 763 \\
Last Letter Concatenation & 350 & 1029 \\ \hline
\end{tabular}%
}
\caption{Statistics of augmented samples by our mentor model (FlanT5-large) per dataset. Here, three CoT rationales per question are generated by our mentor models for augmentation, followed by a filtering process.}
\label{tab:aug_stats}
\end{table}

\section{Effects of Soft Label Distillation} \label{sec:lambda}

\begin{figure}[t]
\centering
    \includegraphics[width=0.48\textwidth]{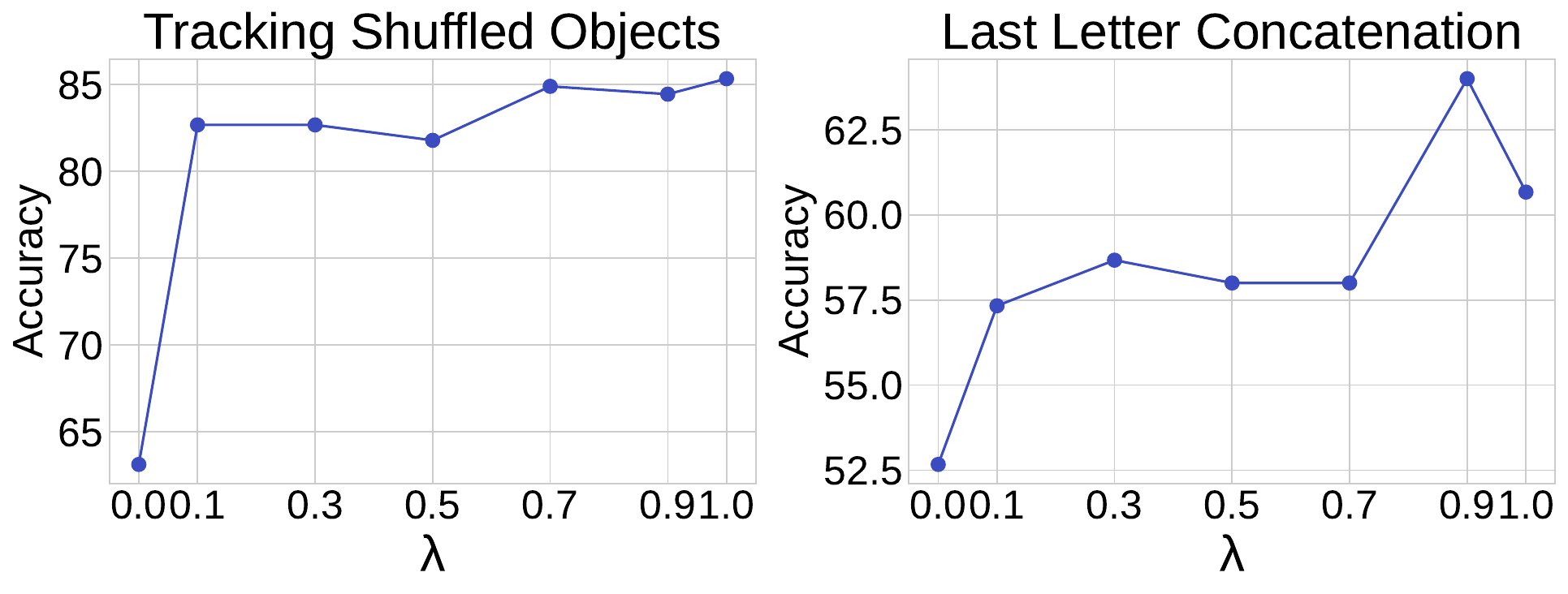}
\caption{Effects of soft label distillation, by varying the value of loss interpolation hyperparameter ($\lambda$).}
\label{fig:lambda}
\end{figure}

In this section, we examine the effects of soft label distillation in Mentor-KD, through differentiating the loss interpolation hyperparamter ($\lambda$) in Equation~\ref{eq:loss_function}. We diversely set the value of $\lambda$ from 0 (no soft labels) to 1 (only soft labels) in this experiment, and set the student model to FlanT5-small using two reasoning tasks.

The results are shown in Figure~\ref{fig:lambda}. We initially observe that the student model's performances are the lowest when no soft labels are introduced to reasoning distillation. However, we also observe that introducing soft labels significantly contribute to performance boosts of the student, implying that the soft labels which mentor models provide are beneficial to student models carrying out multi-step reasoning.
% Furthermore, we notice in Last Letter Concatenation that only using soft labels may leave a negative impact to the student's performance, suggesting that both rationale distillation and soft label distillation must be taken into consideration

\section{API Usage}
As mentioned in Section~\ref{sec:experiment_setup}, we employ GPT-3.5-Turbo as our teacher model throughout all experiments. Access to the model was provided by the OpenAI API. We set the generation temperature to 0.7, following previous works \cite{wang-etal-2023-self_consistency,ho-etal-2023-ftcot}. Our total expenditure for collecting CoT annotations was \$240.5 USD.

\section{Case Study}
In Table~\ref{tab:case_study}, we provide some examples of how our mentor model can successfully augment teacher-incorrect samples on four datasets. Here, we employ GPT-3.5-Turbo as our teacher model, and FlanT5-large as our mentor model. We observe in commonsense reasoning tasks, there are cases where the teacher model does not answer the question faithfully or fails to narrow down its final prediction to a single choice, in contrast to the rationales that the trained mentor model generates. Meanwhile for symbolic reasoning (Last Letter Concatenation), we observe instances where the teacher model makes a final prediction inconsistent of its multi-step rationale, on contrary with the mentor model.

% Inconsistency 관련, it depends 어쩌구 관

% Contribution of Student’s Feedback Additional examples of the LLM’s generated rationales are presented in Table 13. We observe that the teacher LLM, ChatGPT, is capable of generating more detailed and precise reasoning steps when provided with student’s feedback (i.e., wrong solution). These detailed reasoning steps can help address the student’s deficiencies and thereby improve the reasoning performance in the subsequent round of learning. Although both rationales, with and without feedback, are correct, their quality can vary. More precise and customized rationales can help the student better understand its own mistakes, especially coupled with our self-reflection learning, which is beneficial for student’s reasoning learning.

\definecolor{grassgreen}{RGB}{76, 187, 23}
\newcommand{\cmark}{\textcolor{grassgreen}{\scalebox{1.5}{\ding{51}}}} % fatter checkmark
\newcommand{\xmark}{\textcolor{red}{\scalebox{1.5}{\ding{55}}}} % fatter cross

\begin{table*}
\centering
\resizebox{\textwidth}{!}{%
\begin{tabular}{@{}cp{18cm}@{}}
\toprule
\multicolumn{2}{c}{\textbf{(a) StrategyQA}} \\ \midrule
\multicolumn{1}{c|}{Question} & Would Jesus understand the Easter Bunny? \\ \midrule
\multicolumn{1}{c|}{Answer} & No \\ \midrule
\multicolumn{1}{c|}{Teacher} & 1. Jesus lived over 2,000 years ago in the Middle East, so it is highly unlikely that he would have encountered the Easter Bunny as it is known in modern Western culture. 2. The Easter Bunny is a symbol of Easter that has its origins in German folklore and was brought to America by German immigrants in the 18th century. It is not a part of Christian religious tradition. 3. Jesus preached about love, forgiveness, and compassion, so it is possible that he would understand and appreciate the idea of the Easter Bunny as a symbol of joy, renewal, and new life. 4. Ultimately, whether or not Jesus would understand the Easter Bunny is a matter of personal interpretation and belief. Some may argue that Jesus would not have a need to understand or acknowledge the Easter Bunny, as his teachings focus on spiritual matters rather than secular traditions. Others may see the Easter Bunny as a harmless and fun part of Easter celebrations. Therefore, the answer is it depends. \xmark \\ \midrule
\multicolumn{1}{c|}{Mentor} & First, we need to consider the historical context of Easter Bunny. Easter Bunny is a mythical Easter bunny that is associated with the Easter holiday. Jesus was a historical figure who lived during the time of the Roman Empire, during the time of the Roman conquests. Jesus was a man who lived in the first century AD, during the time of the Roman Empire. Therefore, it is highly unlikely that Jesus would have understood the Easter Bunny. $--$ \textgreater \ No \ \cmark \\ \midrule
\multicolumn{2}{l}{\vspace{-1.5em}} \\ \midrule

\multicolumn{2}{c}{\textbf{(b) CommonsenseQA}} \\ \midrule
\multicolumn{1}{c|}{Question} & What does someone usually produce a lot of when getting divorce? Answer Choices: A. freedom B. tears C. richer D. depression E. tissues \\ \midrule
\multicolumn{1}{c|}{Answer} & (B) \\ \midrule
\multicolumn{1}{c|}{Teacher} & First, when getting a divorce, someone may produce a lot of tears due to the emotional strain of the process. This would make option B a likely choice.  Second, the individual may also experience a sense of freedom from a difficult or unhappy marriage, making option A a possible answer.  Third, divorce could potentially lead to financial changes, with the individual becoming either richer or poorer depending on the circumstances. However, it is more common for individuals to experience financial strain during divorce proceedings, making option C less likely.  Finally, the emotional toll of divorce can contribute to feelings of depression, making option D a reasonable choice.  While tissues may be needed to dry the tears produced during a divorce, they are not typically something that someone \"produces\" in large quantities during the process. Therefore, option E is the least likely choice.  In conclusion, the most likely answers are B. tears and D. depression. Therefore, the answer is B and D. \xmark \\ \midrule
\multicolumn{1}{c|}{Mentor} & B. tears When someone is going through a divorce, they are likely to experience emotional turmoil and emotional distress. This can lead to tears as they process the loss and the emotional impact of the situation. $--$\textgreater \ B \cmark \\ \midrule
\multicolumn{2}{l}{\vspace{-1.5em}} \\ \midrule

\multicolumn{2}{c}{\textbf{(c) Tracking Shuffled Objects}} \\ \midrule
\multicolumn{1}{c|}{Question} & Alice, Bob, and Claire are playing a game. At the start of the game, they are each holding a ball: Alice has a red ball, Bob has a pink ball, and Claire has a black ball. As the game progresses, pairs of players trade balls. First, Claire and Alice swap balls. Then, Alice and Bob swap balls. Finally, Alice and Claire swap balls. At the end of the game, Alice has the Which choice is true? Answer choices: (A) red ball, (B) pink ball, (C) black ball. \\ \midrule
\multicolumn{1}{c|}{Answer} & (A) \\ \midrule
\multicolumn{1}{c|}{Teacher} & After the first swap: Alice has a black ball, Bob has a pink ball, Claire has a red ball. After the second swap: Alice has a red ball, Bob has a pink ball, Claire has a black ball. After the third swap: Alice has a black ball, Bob has a pink ball, Claire has a red ball. Therefore, at the end of the game, Alice has the (C) black ball. Therefore, the answer is (C). \xmark \\ \midrule
\multicolumn{1}{c|}{Mentor} &  After the first swap: Alice has a black ball, Bob has a pink ball, Claire has a red ball. After the second swap: Alice has a pink ball, Bob has a black ball, Claire has a red ball. After the third swap: Alice has a red ball, Bob has a black ball, Claire has a pink ball. Therefore, at the end of the game, Alice has the (A) red ball. $--$\textgreater \ (A) \cmark \\ \midrule
\multicolumn{2}{l}{\vspace{-1.5em}} \\ \midrule

\multicolumn{2}{c}{\textbf{(d) Last Letter Concatenation}} \\ \midrule
\multicolumn{1}{c|}{Question} &Take the last letters of each words in ``Dino Toby Abigail Manuela" and concatenate them. \\ \midrule
\multicolumn{1}{c|}{Answer} & oyla \\ \midrule
\multicolumn{1}{c|}{Teacher} &1. Dino - Last letter: o, 2. Toby - Last letter: y, 3. Abigail - Last letter: l, 4. Manuela - Last letter: a. Concatenate: oyl. Therefore, the answer is oyl. \xmark \\ \midrule
\multicolumn{1}{c|}{Mentor} & 1. Dino - o, 2. Toby - y, 3. Abigail - l, 4. Manuela - a. Concatenating the last letters of each word, we get: oyla. $--$\textgreater \ oyla \cmark \\ \bottomrule
\end{tabular}%
}
\caption{Case study of rationale augmentations by the mentor model (FlanT5-large).}
\label{tab:case_study}
\end{table*}

\end{document}